В.В. Кромер

**Глоттохронология и проблемы праязыковой реконструкции**

Лексикостатистическая глоттохронология традиционно используется как для определения времени разделения родственных языков, так и для установления самого факта родства и степени его близости. В данной статье для обозначения различных языковых образований (языков, диалектов и др.) мы будем использовать наиболее общий термин – изолект. Обозначение "язык" мы сохраняем лишь для современных нам образований (например, македонский язык). При принятии основных постулатов глоттохронологии (наиболее последовательно изложенных в [1]), временной отрезок, разделяющий изолект-предок и изолект-потомок, определяется по формуле [10, с. 18]

$$t = \frac{\ln c}{\ln r}, \qquad (1)$$

где $c$ – коэффициент совпадений между основным списком соответствующих словарей, $r$ – коэффициент сохраняемости списка за единицу измерения времени $t$ (согласно М. Сводешу – константа).

Коэффициент сохраняемости $r$ принимает различные значения для 100-словного и 200-словного основных списков [7]. Ввиду возможного непостоянства скорости распада словаря целесообразно положить $\ln r = -1$, и тем самым определить единицу измерения лингвистического времени. При длине основного списка в $N_0 = 100$ слов принципиально невозможно задание коэффициента $c$ с точностью более 1%. Исходя из приближенного равенства [4, с. 89]

$$\ln(1-x) \approx -x, \qquad (2)$$

справедливого при $x \ll 1$, вычисление $t$ целесообразно с точностью до 0,01 определенной выше единицы времени. Эту дольную единицу предлагается назвать «сводешем» в честь основателя глоттохронологии Мориса Сводеша (1909–1967) и присвоить ей обозначение *Свод*. Учитывая, что вместо коэффициента $c < 1$ обычно указывается процент совпадения лексики $C = 100c$, формулу (1) можно переписать в виде

$$L = -100\ln\left(\frac{C}{100}\right), \qquad (3)$$

где $L$ измеряется в сводешах. Представляется целесообразным рассматривать $L$ как расстояние между двумя изолектами. В случае прямого родства между двумя изолектами $L$ является временно́й дистанцией между ними, при синхронном сравнении двух родственных изолектов $L$ – расстояние между ними через общий изолект-предок, т.е. $\frac{L}{2}$ является временем дивергенции (расхождения) родственных языков (известная формула $t = \frac{\ln c}{2 \ln r}$ [10, с. 30]). Если сравниваемые изолекты принадлежат к различным эпохам, вычисление $L$ по проценту совпадения по-прежнему является расстоянием между языками через ближайший общий предок. Чтобы вычислить время дивергенции языков, отсчитываемое назад от современности, необходимо прибавить к $L$ отсчитанное назад от современности время фиксации каждого из рассматриваемых языков ($T_1$ и $T_2$) в сводешах, и сумму разделить на 2:

$$T_d = \frac{-100\left(\ln \frac{C}{100}\right) + T_1 + T_2}{2}. \qquad (4)$$

Соответствующая формула для случая, когда один из сравниваемых языков – современный (т.е. $T_2 = 0$), приведена в [10, с. 31].

Измерение времени в сводешах позволяет производить расчеты лингвистических времен без оглядки на постулат постоянства скорости изменения словаря и возможности варьирования этой скорости в зависимости от языка, жанра и эпохи. При наложении шкалы результатов относительных расчетов на шкалу абсолютного (физического) времени появляется возможность, путем идентификации определенных фиксированных точек шкалы лингвистического и абсолютного времени, произвести увязку обоих шкал путем (в общем случае нелинейного) преобразования (деформации) шкалы лингвистического времени, что позволяет произвести датировку лингвистических событий. Далее в тексте процент совпадения слов основного списка сравниваемых изолектов называется коэффициентом и обозначается за $C$.

На основании результатов сопоставления слов основного списка для $k$ языков получается

$$n = \frac{k^2 - k}{2} = \frac{k(k-1)}{2} \qquad (5)$$

отличающихся значений коэффициента $C$. Путем сравнения отдельных членов матрицы коэффициентов $C$ делаются выводы о наличии или отсутствии диалектных (языковых) цепей (сетей) или суждения о распаде существовавших языковых цепей [10, с. 32–34].

Покажем на нескольких примерах, что матрица коэффициентов содержит информацию о временно́й расстановке отдельных изолектов и синхронной их расстановке в виде изолектной цепи.

Пример 1: Матрица коэффициентов совпадения для 4 языков племен группы сэлиш приведена в работе [10, с. 32] и повторена в виде таблицы 1 в данной работе. Вычисленные по формуле (3) расстояния между парами языков приведены в таблице 2.

Таблица 1

| | Язык | 1 | 2 | 3 | 4 |
|---|---|---|---|---|---|
| 1 | Лилуит | – | 48 | 33 | 25 |
| 2 | Шусвап | 48 | – | 50 | 34 |
| 3 | Окэнагон | 33 | 50 | – | 54 |
| 4 | Колумбия | 25 | 34 | 54 | – |

Таблица 2

| Язык | 1 | 2 | 3 | 4 |
|---|---|---|---|---|
| 1 | – | 73 | 111 | 139 |
| 2 | 73 | – | 69 | 108 |
| 3 | 111 | 69 | – | 62 |
| 4 | 139 | 108 | 62 | – |

Из таблицы 2 следует, что парой языков с минимальным взаимным расстоянием является пара языков 3-4. Если считать, что языки 3 и 4 дивергировали из единой точки, то остается необъясненной неодинаковость расстояний от точек 3 и 4 (впредь точки будут идентифицироваться с отображаемыми на точку изолектами) до любой другой точки (например, 1 или 2). Существующая разница может быть объяснена тем, что языки 3 и 4 дивергируют из разных точек изолектной цепи. Найдем среднее расстояние от точки 3 до остальных точек (т.е. 1 и 2). $L_3 = \dfrac{L_{31}+L_{32}}{2} = \dfrac{111+69}{2} = 90$ (За $L_{mn}$ обозначено расстояние между точками $m$ и $n$). Соответственно $L_4 = \dfrac{L_{41}+L_{42}}{2} = \dfrac{139+108}{2} = 124$ (произведено округление). Разница расстояний $\Delta L_{43} = L_4 - L_3 = 124 - 90 = 34$. Дендрограмма соответствующего участка изображена на рисунке 1*а*. В данной работе на дендрограммах языки отображаются ромбами, узлы дендрограмм (изолекты) – точками. Языки и изолекты отмечены числами, выделенными жирным шрифтом. Числа у звеньев означают их длину в сводешах. Именно расстановка языков 3 и 4 и их изолектов-предков, отображенная на рисунке 1*а*, объясняет расстояние между ними в 62 Свод и разницу расстояний в 34 Свод при связи с внешней по отношению к языкам 3 и 4 языковой системой через узел 3-4. Вводится дополнительное положение о выборе из нескольких возможных вариантов расстановки варианта с наибольшей вертикальной (хронологической) глубиной. Перепишем таблицу 2, заменив строки, соответствующие языкам 3 и 4, строкой 3-4 (для узла 3-4). В соответствующих ячейках таблицы 3 стоят расстояния от узла 3-4 до точек 1 и 2, вычисляемые с учетом положения узла 3-4. В качестве примера:
$L_{(3-4)1} = \dfrac{(L_{31}-14)+(L_{41}-48)}{2} = \dfrac{(111-14)+(139-48)}{2} = \dfrac{97+91}{2} = 94$. Смысл расчета следующий: от $L_{31}$ отнимается 14 (расстояние между точкой 3 и узлом 3-4), от $L_{41}$ отнимается 48 (расстояние между точкой 4 и узлом 3-4).

Полученные два расстояния (97 и 91) являются двумя независимыми оценками расстояния от узла 3-4 до точки 1. Математическим ожиданием этого расстояния является среднее значение двух равновесных оценок, т.е. 94. Согласно таблице 3 наиболее близки точка 2 и узел 3-4. Разница расстояний между новым узлом 3-4 и точкой 2 до единственной внешней точки 1 (через узел 2-4) равна $\Delta L_{(3-4)2} = L_{(3-4)1} - L_{21} = 94 - 73 = 21$. (Для дальнейших расчетов с целью получения целочисленных значений принимаем $\Delta L_{(3-4)2} = 20$). Соответствующий участок дендрограммы приведен на рисунке 1б. Перепишем таблицу 3, заменив строки, соответствующие языку 2 и узлу 3-4, строкой 2-4 (для узла 2-4) (Таблица 4). Расстояние от узла 2-4 до точки 1 вычислено с учетом разных весов точки 2 (представляющей один язык – одно измерение) и узла 3-4 (представляющего два языка – два измерения):

$$L_{(2-4)1} = \frac{(L_{21} - 19) + 2(L_{(3-4)1} - 39)}{3} = \frac{(73 - 19) + 2(94 - 39)}{3} = \frac{54 + 2 \cdot 55}{3} = 55.$$

Таблица 3

| Язык | 1 | 2 | 3-4 |
|---|---|---|---|
| 1 | – | 73 | 94 |
| 2 | 73 | – | 58 |
| 3-4 | 94 | 58 | – |

Таблица 4

| Язык | 1 | 2-4 |
|---|---|---|
| 1 | – | 55 |
| 2-4 | 55 | – |

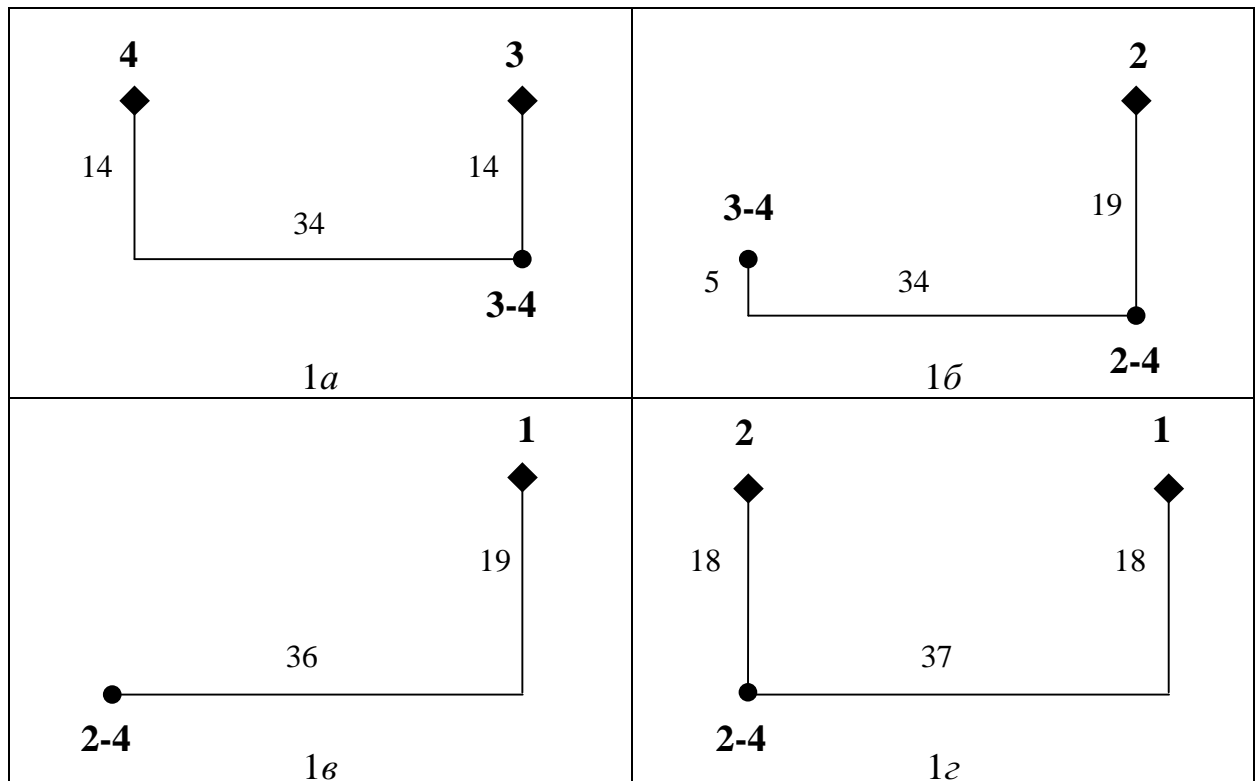

Рис. 1

Расстояние от точки 1 до узла 2-4 составляет 55 Свод. Возможны варианты включения точки 1 в подсистему 2-4, поскольку отсутствуют дополнительные условия для определения формы последнего подключаемого звена. Выскажем предположение, что подключенное звено имеет наиболее простую форму (рисунок 1*в*) и проверим его, начав конструирование дендрограммы с обсчета звена 1-2. Результаты приведены на рисунке 1*г*. Практическое совпадение результатов двух независимых расчетов позволяет считать высказанное предположение справедливым. В то же время нормальным порядком конструирования дендрограммы является описанный ранее, с порядком подключения звеньев, определяемом их длиной, начиная с наиболее короткого звена.

Полная дендрограмма языковой системы 1-2-3-4 приведена на рисунке 2. Из дендрограммы можно сделать вывод, что 19 Свод назад существовала изолектная цепь длиной 70 Свод, три изолекта которой развились в современные языки 1 (лилуит), 2 (шусвап) и 3 (окэнагон). Через 5 Свод с рассматриваемого периода изолектная цепь расширилась на 34 Свод, и крайний член цепи за 14 Свод развился в современный язык 4 (колумбия). Аналогичные выводы о наличии языковой цепи в системе языков племен сэлиш сделаны М. Сводешем исходя из качественных соображений [10, с. 33].

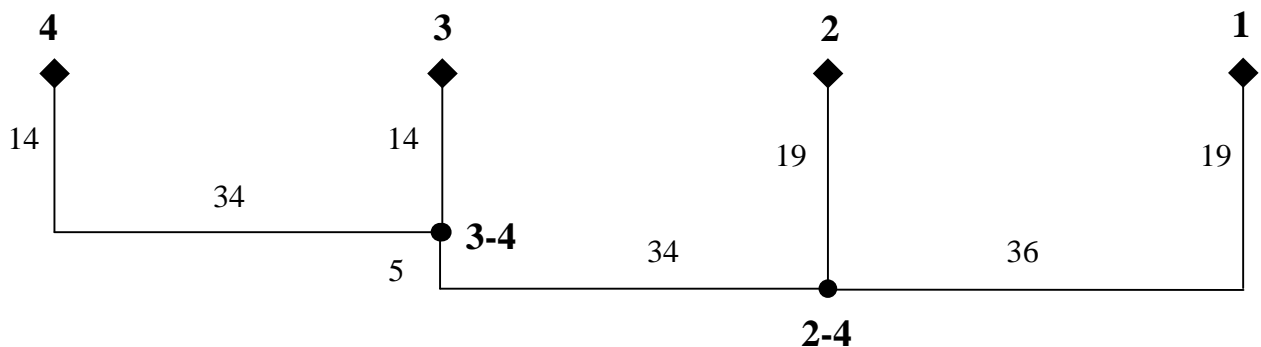

Рис. 2

Посредством располагаемых данных более глубокое проникновение в процесс глоттогенеза невозможно. Для реконструкции более отдаленного от нас состояния языковой системы необходимы данные по внешним по отношению к рассматриваемой системе языкам. Подобную мысль высказывал в своей классической работе М. Сводеш: «Дальнейшее освещение доисторических взаимоотношений между этими языками может быть получено путем изучения взаимоотношений их носителей с соседями, территории которых расположены в других направлениях" [10, с. 33–34]. В частности, учет внешних связей позволяет найти точку связи рассматриваемой системы с внешней системой, а тем самым и наиболее близкий к общему изолекту-предку язык из рассматриваемых четырех.

Точность реконструкции дендрограммы может быть проверена путем измерения в соответствии с рисунком 2 расстояний между современными четырьмя языками. Результаты приведены в таблице 5. Разрешим формулу (3) относительно $C$:

$$C = 100\exp\left(-\frac{L}{100}\right) \qquad (6)$$

и преобразуем данные таблицы 5 в процент совпадения основных списков. Данные приведены в таблице 6.

Таблица 5

| Язык | 1 | 2 | 3 | 4 |
|---|---|---|---|---|
| 1 | – | 74 | 108 | 142 |
| 2 | 74 | – | 72 | 106 |
| 3 | 108 | 72 | – | 62 |
| 4 | 142 | 106 | 62 | – |

Таблица 6

| Язык | 1 | 2 | 3 | 4 |
|---|---|---|---|---|
| 1 | – | 48 | 34 | 24 |
| 2 | 48 | – | 49 | 35 |
| 3 | 34 | 49 | – | 54 |
| 4 | 24 | 35 | 54 | – |

Сравнение данных таблицы 6 (коэффициента $C$ согласно дендрограммы) с данными таблицы 1 (исходные данные) позволяет сделать вывод об адекватности сконструированной дендрограммы эмпирическим данным.

Пример 2: Проведем описанную выше процедуру для другой группы языков сэлиш (также четыре языка). Из этой группы два языка (лилуит и шусвап) общие с рассмотренной выше группой. Сохраним за ними прежние обозначения 1 и 2 соответственно. Языки лоуер фрейзер и нутсак обозначим за языки 5 и 6 соответственно. В таблице 7 приведены исходные данные по проценту совпадения слов основного списка [10, с. 33]. Порядок объединения языков для выявления узлов следующий: 5-6, 1-2, (1-2)-(5-6). Вычисленные расстояния между заданными языками и выявленными узлами приведены в таблице 8. Дендрограммы отдельных участков языковой системы 1-2-5-6 изображены на рисунке 3. Вопрос с последним включаемым в систему звеном (1-2)-(5-6) более сложен. Практическое равенство двух вариантов звена 1-2 в системе 1-2-3-4, рассчитанных при включении его в систему последним (рис. 1*б*, 1*в*) и первым (рис. 1*г*) ведет к единственному варианту дендрограммы системы. Включение звена 1-5 в систему 1-2-5-6 последним (через узлы 1-2 и 5-6) допускает в качестве подсистемы с наименьшей глубиной вариант по рисунку 3*в*, при включении же звена в систему первым и соблюдении постулата о примате варианта с наибольшей хронологической глубиной мы получаем вариант 3*г*. Принять же этот вариант в качестве единственного возможного мы не можем, поскольку при этом нарушается постулат о первоочередности включения в конструируемую дендрограмму наиболее короткого звена. Итак, реализованы могут быть как крайние варианты (с наибольшей шириной изолектной цепи либо с наибольшей диахронической глубиной), так и любой промежуточный.

Как отмечалось выше, выбор окончательного варианта предполагает учет внешних связей, каковыми мы не располагаем.

Таблица 7

| Язык | 2 | 1 | 5 | 6 |
|---|---|---|---|---|
| 2 | – | 48 | 19 | 19 |
| 1 | 48 | – | 28 | 26 |
| 5 | 19 | 28 | – | 58 |
| 6 | 19 | 26 | 58 | – |

Таблица 8

| Язык | 2 | 1 | 5 | 6 | 5-6 | 1-2 |
|---|---|---|---|---|---|---|
| 2 | – | 73 | 166 | 166 | 139 | 104 |
| 1 | 73 | – | 127 | 135 | * | * |
| 5 | 166 | 127 | – | 54 | * | * |
| 6 | 166 | 135 | 54 | – | * | * |
| 5-6 | 139 | 104 | * | * | – | 85 |
| 1-2 | * | * | * | * | 85 | – |

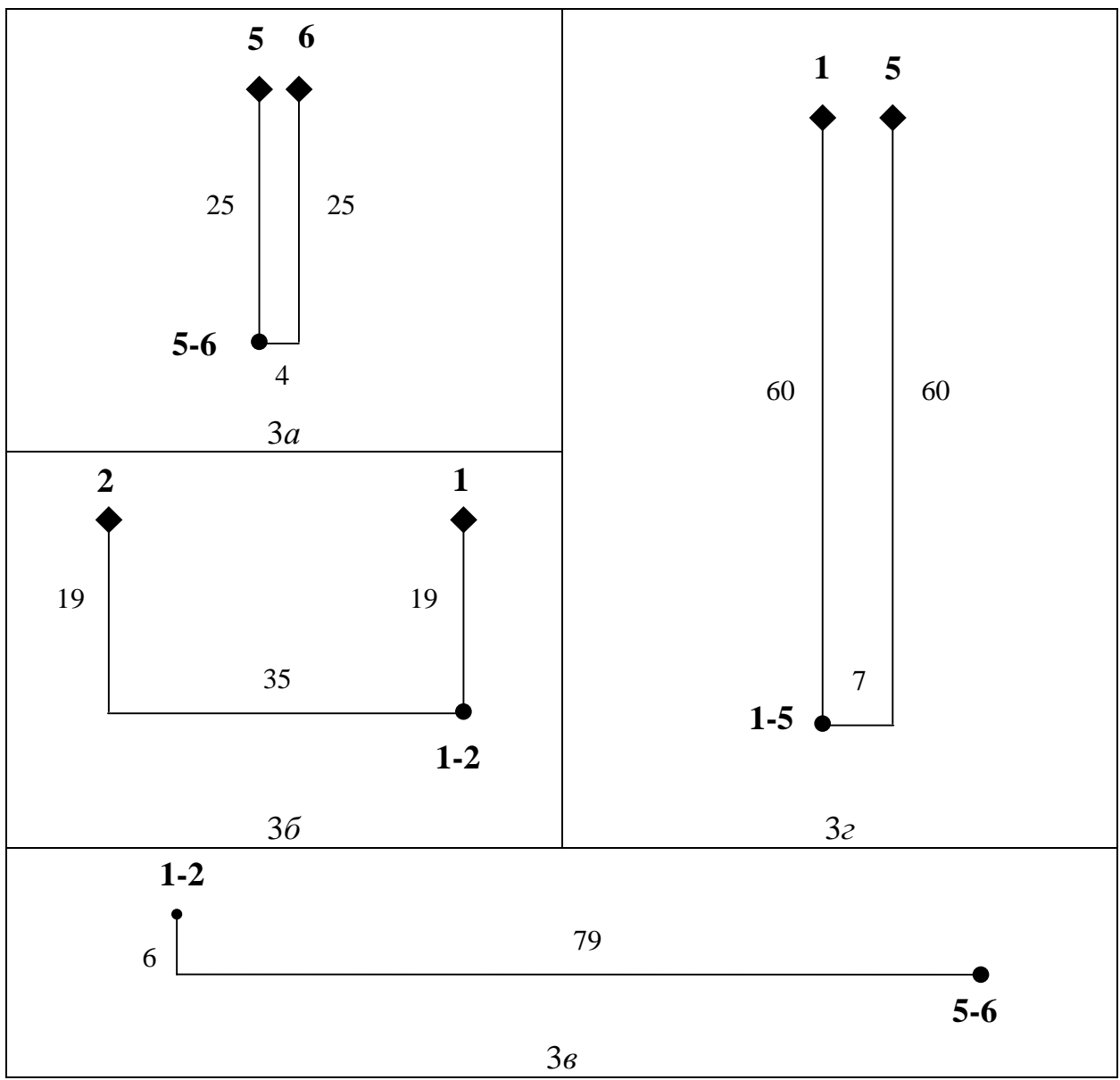

Рис. 3

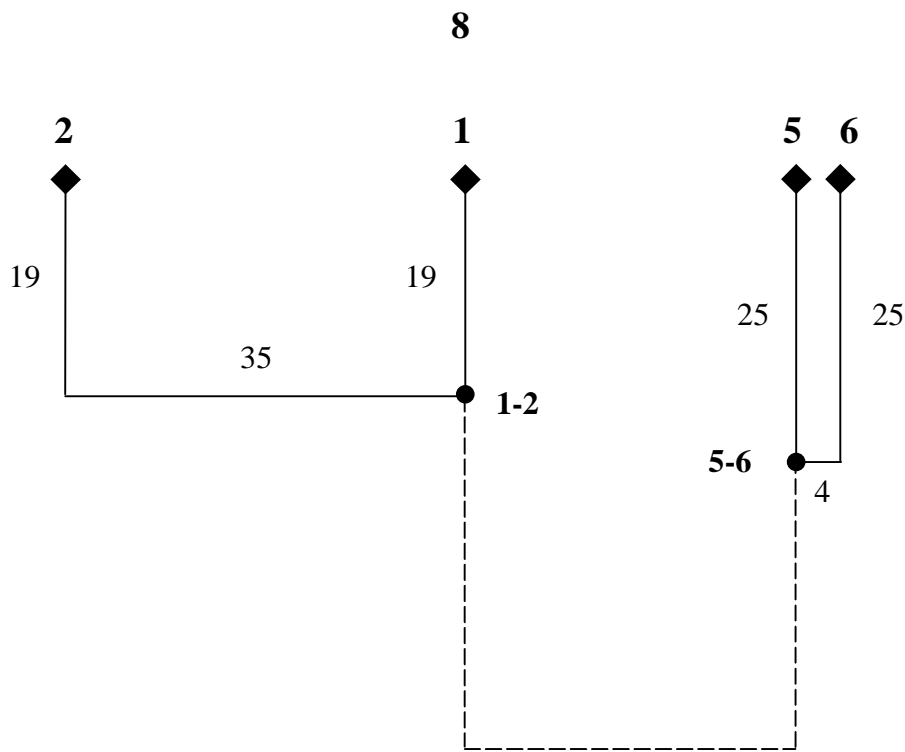

Рис. 4

Полная дендрограмма системы изображена на рисунке 4. Не определяемое однозначно в рамках представленных данных звено дендрограммы изображено штриховой линией. Известна лишь общая длина участка между узлами 1-2 и 5-6, равная 85 Свод. М. Сводеш на основании качественного анализа приведенных данных делает вывод о распаде существовавшей старой диалектной цепи, т.е. склоняется в пользу варианта 3*г*.

При сравнении частных дендрограмм совпадающих участков двух систем (связанных общими для двух групп языками 1 и 2) по рисункам 1*в* и 3*б* выявляется их полное (в пределах точности расчетов) совпадение. Данный результат не является тривиальным, поскольку общность языков гарантирует лишь равенство расстояний между ними (согласно исходным данным таблиц 1 и 7), форма же частной дендрограммы определяется всем набором данных таблиц 1 и 7. Проиллюстрируем сказанное: произвольное изменение коэффициента совпадения между языками 1 и 4 на $+4\%$ или $-4\%$ ведет к частным дендрограммам соответственно рисунков 5*а* и 5*б*, отличающимся как друг от друга, так и от дендрограммы рисунка 1*в*, что иллюстрирует высокую внутреннюю согласованность исходных данных и чувствительность метода.

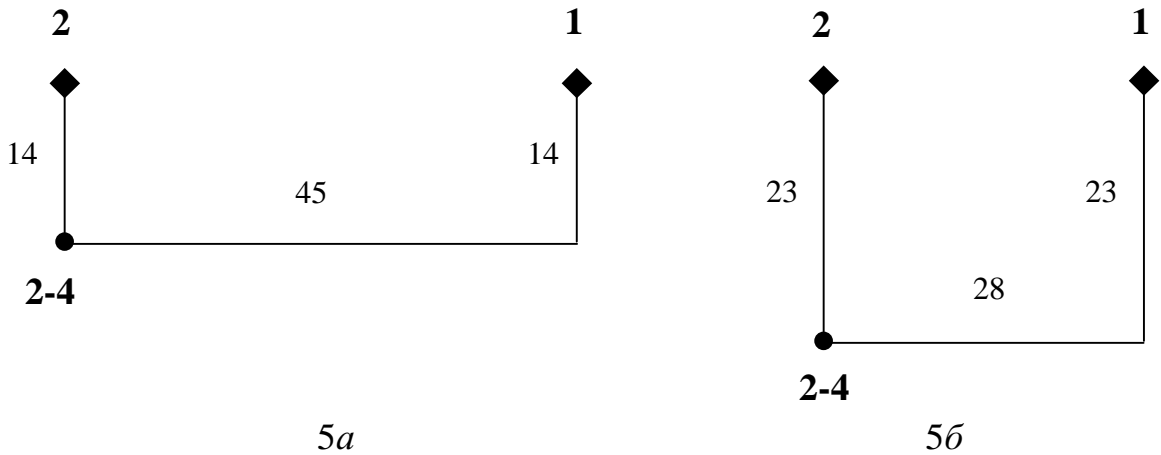

*5а*                        *5б*

Рис. 5

Наличие общих точек на дендрограммах систем 1-2-3-4 и 1-2-5-6 позволяет произвести совмещение двух дендрограмм и конструирование единой дендрограммы системы 1-2-3-4-5-6 (рисунок 6), а также определить отсутствующие в исходных данных [10, с. 32–33] предполагаемые проценты совпадения слов основного списка языков 3 и 4, с одной стороны, и 5 и 6, с другой стороны, для чего вначале подсчитываем в соответствии с дендрограммой рис. 6 расстояния между языками, а затем по формуле (6) – процент совпадений. Данные по расстояниям сведены в таблицу 9, а по проценту совпадений – в таблицу 10.

Таблица 9                   Таблица 10

| Языки | | 5 | 6 |
|---|---|---|---|
| | | лоуер фрейзер | нутсак |
| 3 | окэнагон | 199 | 203 |
| 4 | колумбия | 233 | 237 |

| Языки | лоуер фрейзер | нутсак |
|---|---|---|
| окэнагон | 14 | 13 |
| колумбия | 10 | 9 |

Принято считать, что при количестве общей лексики менее 15–20% метод глоттохронологии не дает надежных результатов, и доказать родство языков фактически невозможно [6]. Коэффициент совпадений между крайними языками выявленной языковой цепи (колумбия и нутсак) предполагается равным 9%, т.е. гораздо меньшим граничного значения в 15–20%, однако надежное выявление языковой цепи и установление промежуточных звеньев позволяет говорить о родстве.

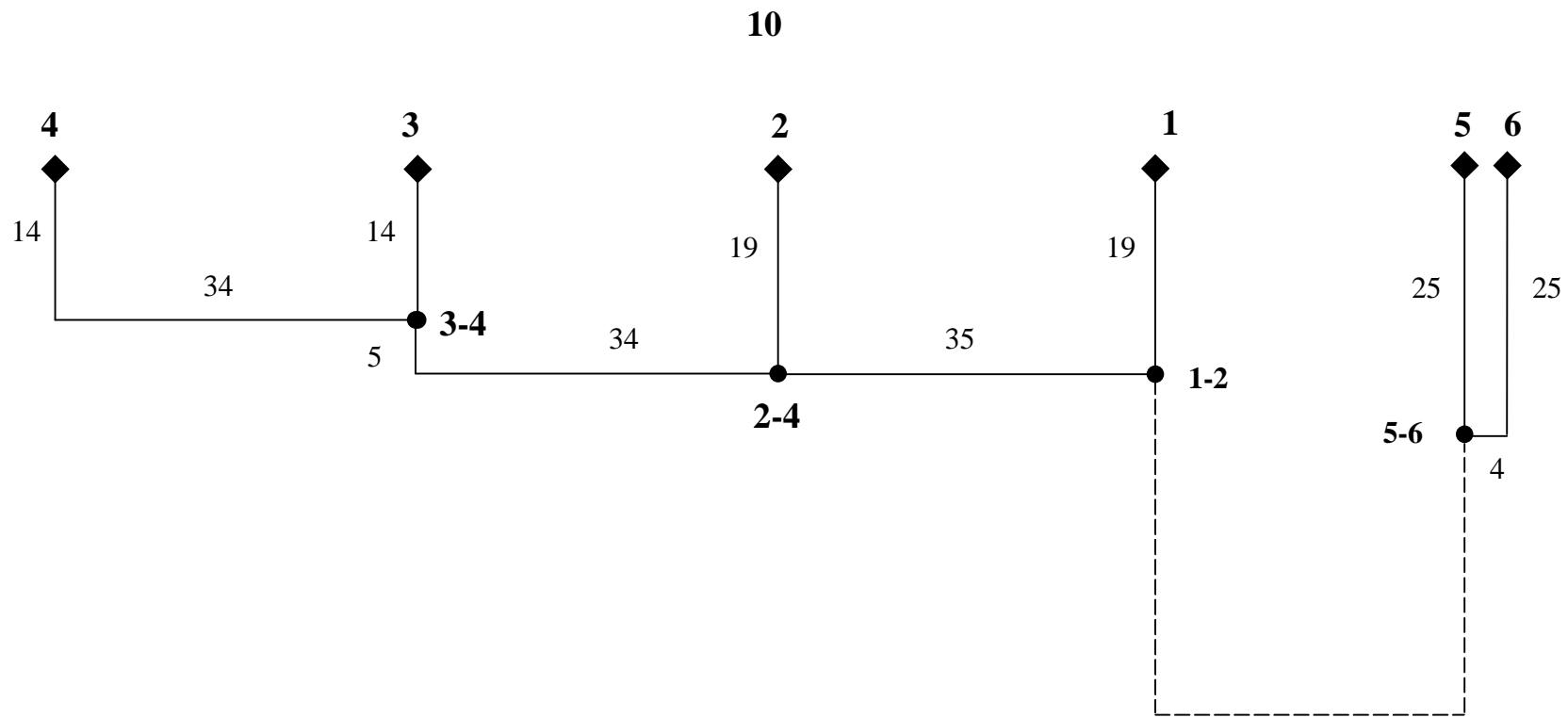

Рис. 6

Пример 3: Применим описанную методику для построения дендрограммы балтославянских языков. Исходные данные по проценту совпадения слов основного списка содержатся в [13] (цитируется по [6]) и приведены в таблице 11. (Нумерация языков в примере 3 не имеет отношения к нумерации языков в примерах 1 и 2).

Таблица 11

| | Язык | 1 | 2 | 3 | 4 | 5 | 6 | 7 | 8 | 9 | 10 | 11 | 12 | 13 | 14 | 15 |
|---|---|---|---|---|---|---|---|---|---|---|---|---|---|---|---|---|
| 1 | Литовский | – | 68 | 49 | 47 | 47 | 47 | 43 | 44 | 46 | 46 | 46 | 44 | 44 | 45 | 46 |
| 2 | Латышский | 68 | – | 44 | 45 | 40 | 44 | 40 | 41 | 42 | 43 | 45 | 40 | 41 | 41 | 42 |
| 3 | Прусский | 49 | 44 | – | 41 | 41 | 40 | 39 | 42 | 42 | 42 | 42 | 40 | 41 | 39 | 40 |
| 4 | Русский | 47 | 45 | 41 | – | 86 | 92 | 77 | 74 | 74 | 73 | 74 | 74 | 71 | 70 | 74 |
| 5 | Украинский | 47 | 40 | 41 | 86 | – | 92 | 76 | 73 | 76 | 74 | 74 | 71 | 73 | 71 | 72 |
| 6 | Белорусский | 47 | 44 | 40 | 92 | 92 | – | 80 | 77 | 80 | 78 | 78 | 76 | 77 | 74 | 77 |
| 7 | Польский | 43 | 40 | 39 | 77 | 76 | 80 | – | 81 | 85 | 83 | 80 | 79 | 75 | 71 | 74 |
| 8 | Чешский | 44 | 41 | 42 | 74 | 73 | 77 | 81 | – | 92 | 87 | 87 | 84 | 79 | 75 | 74 |
| 9 | Словацкий | 46 | 42 | 42 | 74 | 76 | 80 | 85 | 92 | – | 87 | 86 | 80 | 80 | 76 | 75 |
| 10 | Н.-лужицкий | 46 | 43 | 42 | 73 | 74 | 78 | 83 | 87 | 87 | – | 94 | 78 | 74 | 73 | 71 |
| 11 | В.-лужицкий | 46 | 45 | 42 | 74 | 74 | 78 | 80 | 87 | 86 | 94 | – | 78 | 77 | 76 | 73 |
| 12 | Словенский | 44 | 40 | 40 | 74 | 71 | 76 | 79 | 84 | 80 | 78 | 78 | – | 85 | 75 | 76 |
| 13 | Сербский | 44 | 41 | 41 | 71 | 73 | 77 | 75 | 79 | 80 | 74 | 77 | 85 | – | 84 | 80 |
| 14 | Македонский | 45 | 41 | 39 | 70 | 71 | 74 | 71 | 75 | 76 | 73 | 76 | 75 | 84 | – | 86 |
| 15 | Болгарский | 46 | 42 | 40 | 74 | 72 | 77 | 74 | 74 | 75 | 71 | 73 | 76 | 80 | 86 | – |

Контрольный обсчет полученного первого варианта дендрограммы показывает, что при верном в среднем (что гарантируется методом) восстановлении исходных данных таблицы 11, каждому языку свойственна своя погрешность, которую мы будем характеризовать дисперсией восстановленных по дендрограмме расстояний между языками. (Размеры звеньев дендрограммы, в отличие от процента совпадений, аддитивны и к ним применимы процедуры математической статистики). Не выясняя на данном этапе причин этой погрешности, которые могут быть как причинами лингвистического характера, так и быть экстралингвистическими (неполнотой данных, промахами при измерениях, типографскими опечатками), найдем возможность обойти источник погрешности при оптимальном использовании всей доступной информации.

Для построения второго варианта дендрограммы присвоим каждому языку «вес», с точностью до произвольного коэффициента пропорциональности равный обратной дисперсии восстановленных данных по расстояниям между языками. Присвоенные языкам веса приведены в таблице 12.

Расчет расстояний до узлов и разницы расстояний для конструирования дендрограммы подсчитываются с учетом весов языков, что обеспечивает приоритет наиболее надежным данным. По второму варианту дендрограммы вновь посчитана погрешность восстановления данных и уточнены веса. По уточненным весам (незначительно отличающимся от первона-чально присвоенных) рассчитан третий (окончательный) вариант дендрограммы (рисунок 7). Дальнейшее уточнение весов и перерасчет дендрограммы не имеют смысла, поскольку даже между первым и третьим вариантами дендрограммы расхождения несущественны, а между вторым и третьим практически отсутствуют.

Таблица 12

| № | Язык | Вес | № | Язык | Вес | № | Язык | Вес |
|---|---|---|---|---|---|---|---|---|
| 1 | Литовский | 8 | 6 | Белорусский | 16 | 11 | В.-лужицкий | 11 |
| 2 | Латышский | 5 | 7 | Польский | 10 | 12 | Словенский | 4 |
| 3 | Прусский | 13 | 8 | Чешский | 9 | 13 | Сербский | 3 |
| 4 | Русский | 5 | 9 | Словацкий | 21 | 14 | Македонский | 7 |
| 5 | Украинский | 10 | 10 | Н.-лужицкий | 13 | 15 | Болгарский | 16 |

Все предыдущие дендрограммы строились с учетом единого масштаба по вертикали и горизонтали. Особенности дендрограммы балтославянских языков (редкость лингвистических событий в далеком прошлом и насыщенность событиями сравнительно недавнего прошлого, малочисленность балтийских и сравнительная многочисленность славянских языков) заставляют отказаться от соблюдения масштаба при ее представлении. Размеры звеньев дендрограммы по-прежнему отмечаются числами при соответствующих отрезках.

Анализ дендрограммы рисунка 7 выявляет существование около 30 Свод тому назад балтославянской изолектной цепи шириной около 25 Свод и разделение ее на ветви, соответствующие славянским, восточно-балтийским (собственно балтийским) и прусскому языкам, восточно-балтийские языки занимают при этом промежуточное положение между славянскими и прусским языками. 15 Свод тому назад происходит разделение единого восточно-балтийского языка на литовский и латышский, а 11 Свод тому назад – отдельные изолекты изолектной цепи шириной в 6–8 Свод дали начало западной и восточной группам и восточной подгруппе южной группы славянских языков. К западной группе относятся польский, чешский, словацкий, нижнелужицкий и верхнелужицкий языки, к восточной – русский, украинский и белорусский, а к восточной подгруппе южной группы – македонский и болгарский [3, 5]. Отнести в пределах дендрограммы словенский и сербский языки, традиционно относимые к западной подгруппе южной группы славянских языков [5], к какой-либо из вышеназванных групп (подгрупп), не представляется возможным ввиду их низкого веса согласно таблице 12 и вследствие этого «размытости» их положения на дендрограмме как относительно друг друга, так и относительно других балтославянских языков.

**13**

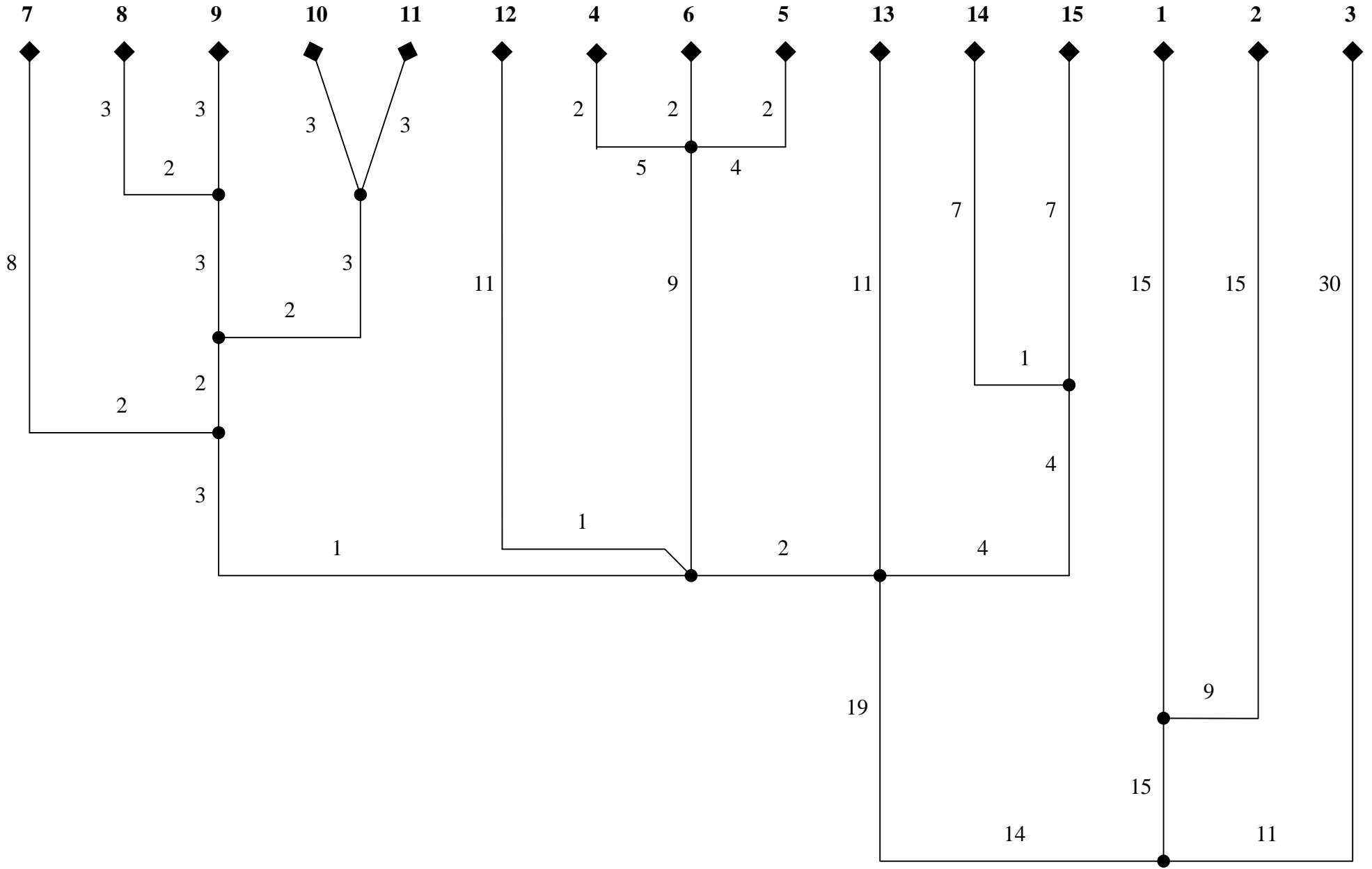

Как уже отмечалось выше, учет лишь внутрисистемных связей не позволяет установить точку связи системы с внешним окружением (для чего необходим учет межсистемных связей, например с языками романской или германской языковых групп). При параметризации подобных связей с целью взаимной увязки отдельных дендрограмм (по группам языков) и увязки их в единую дендрограмму языковой семьи, целесообразно выделение в пределах каждой группы трех-четырех языков, восходящих непосредственно к изолекту-предку данной группы (или подсемьи). Из рисунка 7 с учетом возможной ошибки в определении ширины изолектной цепи следует, что из восточно-балтийских языков таким требованиям отвечает литовский язык, из западнославянских языков, скорей всего, словацкий язык, из восточнославянских языков – белорусский язык. Следует отметить также, что названным языкам (литовскому, словацкому и белорусскому) согласно таблице 12 присвоены наибольшие веса в своих группах. В то же время их «выделенность» на дендрограмме не является следствием их веса (т.е. артефактом), поскольку это выделенное положение наблюдается и при конструировании дендрограммы без учета весов (т.е. при равновесности языков).

При анализе дендрограмм, подобных изображенным на рис. 6 и 7, встает вопрос о лингвистической интерпретации отрезков дендрограмм. Вертикальные отрезки являются линиями диахронного дивергентного развития соответствующего изолекта и отражают закономерный процесс смены лексического состава языка (операционализируемый в рамках данной работы сменой слов основного списка по М. Сводешу), определяемый ограниченной репродуктивной способностью слова в целом (и конечной длительностью его жизни), предопределенной, в свою очередь, активностью и стабильностью исходного значения слова [9]. Суперпозиция таких частных жизненных циклов слов, входящих в основной список, и задает общую динамику вертикального развития языка. Горизонтальные отрезки являются звеньями синхронной изолектной цепи. Ширина звена (величина бокового сдвига) $L$ в сводешах ввиду справедливости формулы (2) при малых $L$ примерно равна количеству замещенных слов основного 100-словного списка (например, в результате языковой конвергенции – скрещивания [11]), а при бóльших $L$ равна $(100-C)$, где $C$ вычисляется по формуле (6).

В таблице 13 в порядке увеличения приведены дисперсии восстановленных по дендрограмме рисунка 7 значений расстояний между языками. Высказывается предположение, что языки с малыми дисперсиями (в данном примере словацкий, болгарский, нижнелужицкий, белорусский) – это языки, в процессе своего развития либо не подвергавшиеся конвергенции, либо конвергировавшие с неродственными языками (языками с отдаленным родством). Языки с большой дисперсией (латышский, словенский, сербский) конвергировали с близкородственными языками.

Таблица 13

| № | Язык | Дисп. | № | Язык | Дисп. | № | Язык | Дисп. |
|---|---|---|---|---|---|---|---|---|
| 9 | Словацкий | 2 | 5 | Украинский | 7 | 1 | Литовский | 13 |
| 15 | Болгарский | 3 | 7 | Польский | 8 | 4 | Русский | 14 |
| 10 | Н.-лужицкий | 4 | 8 | Чешский | 9 | 2 | Латышский | 21 |
| 6 | Белорусский | 5 | 3 | Прусский | 10 | 12 | Словенский | 22 |
| 11 | В.-лужицкий | 7 | 14 | Македонский | 11 | 13 | Сербский | 27 |

На дендрограмме боковой сдвиг идеализированно изображен как единовременное событие, однако инертность лингвистических процессов предполагает развитие процесса лексического замещения во времени, т.е. единство языкового развития по вертикали и горизонтали. Так, развитие болгарского языка согласно дендрограмме рисунка 7 может быть представлено вариантами, некоторые из которых изображены на рисунке 8.

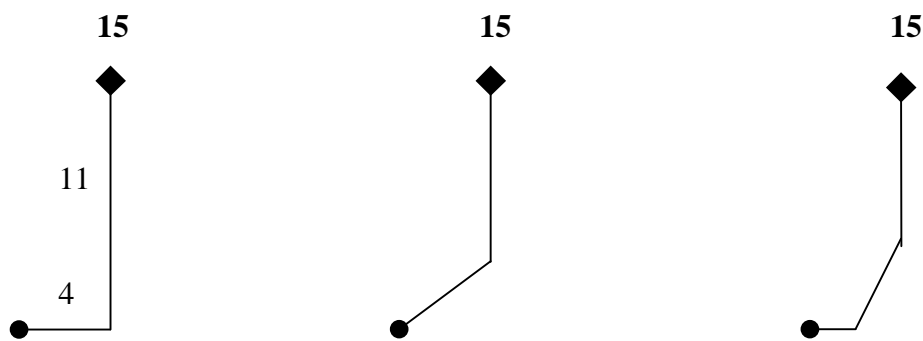

Рис. 8

Привлечение к построению дендрограммы дополнительного языкового материала приводит к более детальной проработке дендрограммы и наоборот – исключение части материала упрощает дендрограмму и огрубляет представление процесса в развитии. В качестве примера произведем построение дендрограммы на основе данных таблицы 9, выбрав из нее украинский, словацкий, нижнелужицкий и болгарский языки. Дендрограмма изображена на рисунке 9. Сравнение ее с дендрограммой рисунка 7 позволяет сделать вывод, что сохранены общие черты дендрограммы рисунка 7 в части выбранных языков с потерей частных особенностей либо их огрублением. Потеря частных особенностей определяется формулой (5), из которой следует примерно квадратичная зависимость информационной емкости дендрограммы от количества сравниваемых языков, т.е. качество проработки дендрограммы как по горизонтали, так и по вертикали пропорционально количеству языков.

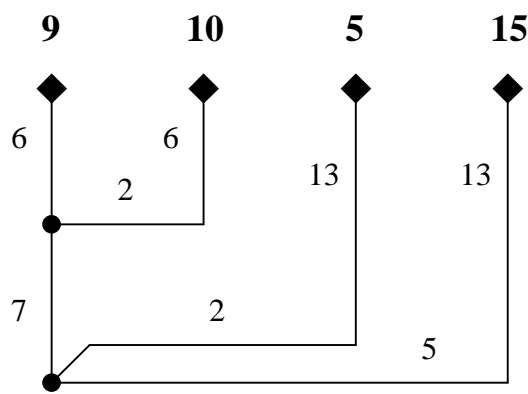

Рис. 9

При увеличении степени проработки дендрограммы образования, традиционно называемые «праязыками», «размазываются» по вертикали и горизонтали. Возможно, что нечто подобное имел в виду русский компаративист В.К. Поржезинский, считая, что праязык реален при учете его хронологических и диалектных рамок [8], что в нашей модели соответствует вертикальному и горизонтальному направлениям на дендрограмме. В упрощенном и схематичном виде вскрытые в данной работе соотношения на дендрограммах между современными языками и изолектами-предками изображены на схеме 2′ работы [2, с. 45].

Согласно таблице 13 наибольшая погрешность восстановления данных присуща словенскому и сербскому языкам. Эта же пара языков характеризуется значительной разностью между измеренным и восстановленным по дендрограмме значениями – соответственно 16 и 25 Свод. (Превышается эта разница лишь парами языков «русский язык – латышский язык» и «сербский язык – македонский язык»). Для всех этих трех пар языков характерна тенденция значительного превышения восстановленного по дендрограмме расстояния над измеренным. Возникает впечатление, что близкое расстояние между современными словенским и сербским языками – результат взаимной конвергенции в недалеком прошлом. Возможно, здесь дендрограмма улавливает взаимную конвергенцию и элиминирует ее до некоторой степени, «разводя» словенский и сербский языки и помещая на дендрограмме словенский язык ближе к западнославянским и восточнославянским языкам, а сербский язык – на полпути между названными языковыми группами и восточной подгруппой южнославянских языков. Во всяком случае, дендрограмма не дает оснований говорить о том, что примерно 11 Свод тому назад при разделении диалектной цепи единого славянского языка на западнославянский, восточнославянский и восточно-южнославянский языки подгруппа т.н. западно-южнославянских языков развивалась на основе одного диалекта.

**Библиографический список**

1. Арапов М.В., Херц М.М. Математические методы в исторической лингвистике. М., 1974.
2. Беликов В.И. Древнейшая история и реальность лингвогенетических дендрограмм // Лингвистическая реконструкция и древнейшая история Востока (Материалы к дискуссиям международной конференции). Т. 1. М.: Наука, 1989. С. 44–54.
3. Бернштейн С.Б. Славянские языки // Языкознание. Большой Энциклопедический словарь. М.: Большая Российская Энциклопедия, 1998. С. 460–461.
4. Бронштейн И.Н., Семендяев К.А. Справочник по математике. М.: Наука, 1986.
5. Гудков В.П. Южнославянские языки // Языкознание. Большой Энциклопедический словарь. М.: Большая Российская Энциклопедия, 1998. С. 600–601.
6. Дьячок М.Т. Глоттохронология: пятьдесят лет спустя // Сибирский лингвистический семинар. 2002. № 1. Новосибирск.
7. Иванов В.С. Глоттохронология // Языкознание. Большой Энциклопедический словарь. М.: Большая Российская Энциклопедия, 1998. С. 109–110.
8. Нерознак В.П. Праязыки: реконструкт или реальность // Сравнительно-историческое изучение языков разных семей. Теория лингвистической реконструкции. М., 1988. С. 26–43.
9. Поликарпов А.А. Циклические процессы в становлении лексической системы языка: моделирование и эксперимент. Автореф. дис. … д-ра филол. наук. М., 1998.
10. Сводеш М. Лексикостатистическое датирование доисторических этнических контактов (на материале племен эскимосов и североамериканских индейцев) // Зарубежная лингвистика. I (общ. ред. В.А. Звегинцева и Н.С. Чемоданова). М.: Прогресс, 1999.
11. Старостин С.А. Скрещивание языков // Языкознание. Большой Энциклопедический словарь. М.: Большая Российская Энциклопедия, 1998. С. 457.
12. Старостин С.А. Сравнительно-историческое языкознание и лексикостатистика // Лингвистическая реконструкция и древнейшая история Востока (Материалы к дискуссиям международной конференции). Т. 1. М.: Наука, 1989. С. 3–39.
13. Girdenis A., Maziulis V. Baltu kalbu divercencine chronologija // Baltistica. T. XXVII (2). Vilnius, 1994. P. 9.